\def\year{2018}\relax
\documentclass[letterpaper]{article} 
\usepackage{aaai18}  
\usepackage{times}  
\usepackage{helvet}  
\usepackage{courier}  
\usepackage{url}  
\usepackage{graphicx}  
\newcommand{\citet}[1]
{\citeauthor{#1} \shortcite{#1}}
\newcommand{\citep}{\cite}

\usepackage[table,svgnames]{xcolor}
  \definecolor{diffstart}{named}{Blue}
  \definecolor{diffincl}{named}{Green}
  \definecolor{diffrem}{named}{Red}

\newcommand{\coloredComment}[2]
 	{\textcolor{#1}{\begin{quote}
 		
 		\small\textit{#2}
 	\end{quote}}}

\newcommand{\inlinedComment}[2]
	{\textcolor{#1}{\small\textit{#2}}}

\usepackage{caption}
\begin{document}
%
\title{Cross-Language Learning for Program Classification \\ using Bilateral Tree-Based Convolutional Neural Networks}
\author{Nghi D. Q. Bui\\
School of Information Systems \\ Singapore Management University \\ dqnbui.2016@phdis.smu.edu.sg
\And Lingxiao Jiang\\
School of Information Systems \\ Singapore Management University \\ lxjiang@smu.edu.sg
\And Yijun Yu\\
Centre for Research in Computing\\ The Open University, UK\\ yijun.yu@open.ac.uk
}
\maketitle
\begin{abstract}
Towards the vision of translating code that implements an algorithm from one programming language into another, this paper proposes an approach for automated program classification using {\em bilateral tree-based convolutional neural networks} (BiTBCNNs). It is layered on top of two tree-based convolutional neural networks (TBCNNs), each of which recognizes the algorithm of code written in an individual programming language. The combination layer of the networks recognizes the similarities and differences among code in different programming languages. The BiTBCNNs are trained using the source code in different languages but known to implement the same algorithms and/or functionalities. For a preliminary evaluation, 
we use 3591 Java and 3534 C++ code snippets from 6 algorithms we crawled systematically from GitHub.
We obtained over 90\% accuracy 
in the cross-language binary classification task to tell whether any given two code snippets implement a same algorithm.
Also, for the algorithm classification task, i.e., to predict which one of the
six algorithm labels is implemented by an arbitrary C++ code snippet, we
achieved over 80\%  precision.  

\end{abstract}

\section{1. Introduction}
\label{sec:intro}
Software engineers need to classify a code snippet against known algorithms, such as Quick Sort, in order to understand it. All algorithms, however, can be implemented in different programming languages, making it hard to recognise an algorithm from the knowledge of its implementation in other languages. It is, therefore, useful to recognise certain algorithms from programs in different  languages, i.e., performing {\em cross-language program classification}.


For a similar problem of language migration, statistical language models have been studied for tokens~\cite{Nguyen2013}, phrases~\cite{Nguyen2015,Nguyen2016}, or APIs  appeared in the code\cite{Zhong2010,Zhong2013,Nguyen2014,Nguyen2014a,Phan2017}. Some of these (i.e., for language recognition and API migration) have been helped by deep neural networks~\cite{Gu2017,Gu2016}, however, little has been done on cross-language program classification.

This paper proposes to use bilateral neural networks (BiNNs), a technique originally developed for comparing natural language sentences, to recognise code snippets in languages that have similar syntax and potentially semantics. Our basic idea is to construct individual subnetworks to encode abstract syntax trees (ASTs) of individual languages, and then construct a combination layer of subnetworks to encode similarities and differences among code structures in different languages.

Our proposed BiTBCNNs are a combination of three major constructs:
(i) BiNNs using {\em softmax} operation for structured data to be compared for classification;
(ii) a variant of tree-based convolutional neural networks (TBCNNs) on each side of the BiNNs to encode AST structures, independent of the programming language of choice; and
(iii) a unified encoding of AST in multiple programming languages that enables cross-language program classification.
With collected programs in different languages, evaluation has shown that our BiTBCNNs have over 80\% accuracy in classifying them according to their underlying algorithms.


The remainder of the paper is organised as follows: Section 2 presents an overview of the proposed process of classifying code in different programming languages; Section 3 details how we make TBCNNs bilateral for the cross-language program classification tasks; Section 4 evaluates the effectiveness of cross-language benchmarks we collected and how they behave when transferring the models across languages or across algorithms; Section 5 presents related work; and finally, Section 6 concludes the findings and suggests some further directions.

\section{2. Overview of Our Approach}\label{sec:process}
\begin{figure*}[t!]\centering
	\includegraphics[width=0.9\textwidth]{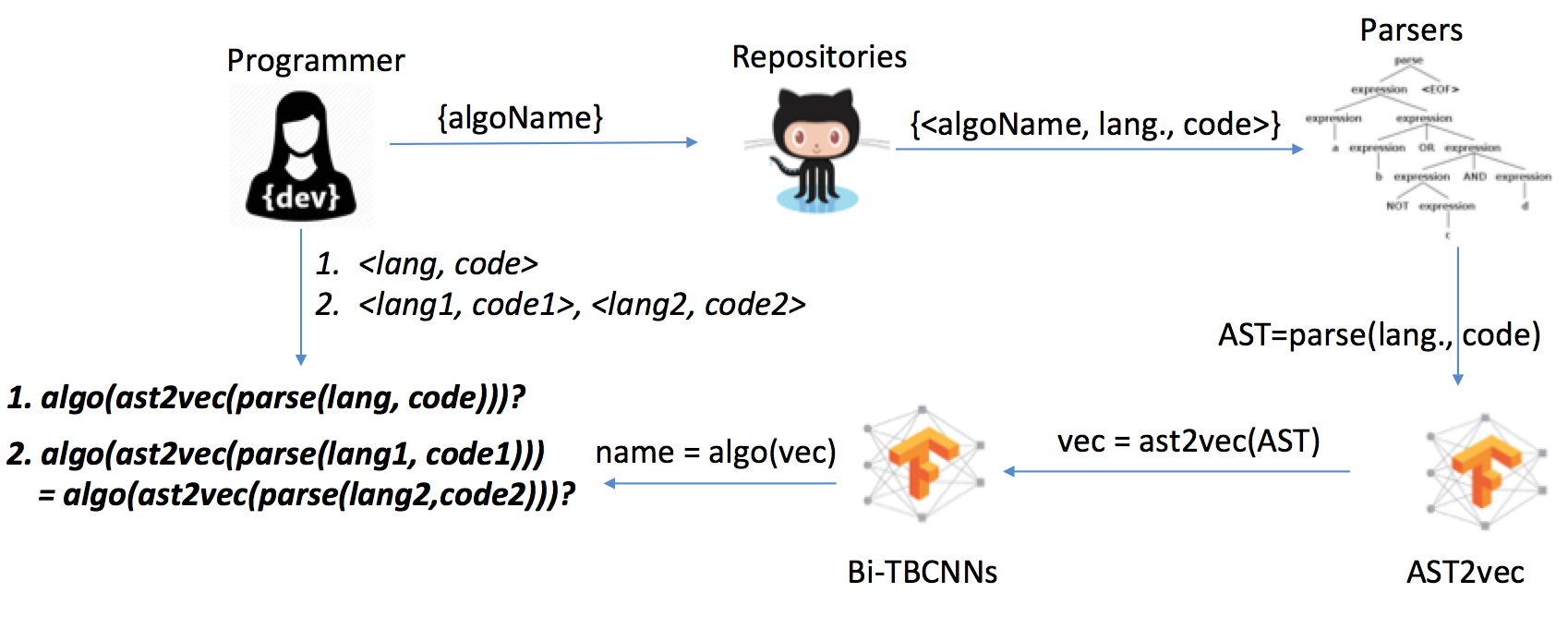}	
	\caption{An overview of our proposed program classification process}
	\label{fig:process}
\end{figure*}

An overview of our approach is illustrated in Figure~\ref{fig:process}. As shown by our open-source repository\footnote{https://github.com/yijunyu/bi-tbcnn}, each step in the process is supported by a simplified {\tt docker} command.

First, a programmer may define a list of algorithms (e.g., mergesort, quicksort, breath-first-search, linkedlist, bublesort, knapsack), and a list of programming languages (e.g., Java, C++, Python).
Given the two lists, automated calls to GitHub RESTful APIs {\em crawl} GitHub repositories to retrieve about 600 instances of program code that implement each algorithm in each programming language.

These code snippets are then {\em parsed}\footnote{https://bitbucket.org/yijunyu/fast} into abstract syntax trees (ASTs) represented in Pickle format.
These ASTs are loaded into memory and converted into a vector form that preserve the distances of similar features ({\em AST2vec}) by training the embeddings of leaf-level tokens, using Tensorflow, according to the programming language (e.g., 385 token types for Java/C/C++/C\#/Objective C in SrcML grammar and 82 token types for Python).

The vectors of both programs of cross-language nature are then combined into {\em BiTBCNNs}, using Tensorflow again, approximating a function that classifies these vectors to algorithm names.

The trained model can be used to answer many types of queries, e.g., (i) what is the algorithm of a piece of code and (ii) whether two programs in different languages implement the same algorithm.

\section{3. Construction of BiTBCNNs}\label{sec:approach}
Our BiTBCNNs construct extends TBCNNs with pretrained vectors as bilateral NNs for cross-language classification.

\begin{figure}[t!]\centering
	\includegraphics[width=0.5\textwidth]{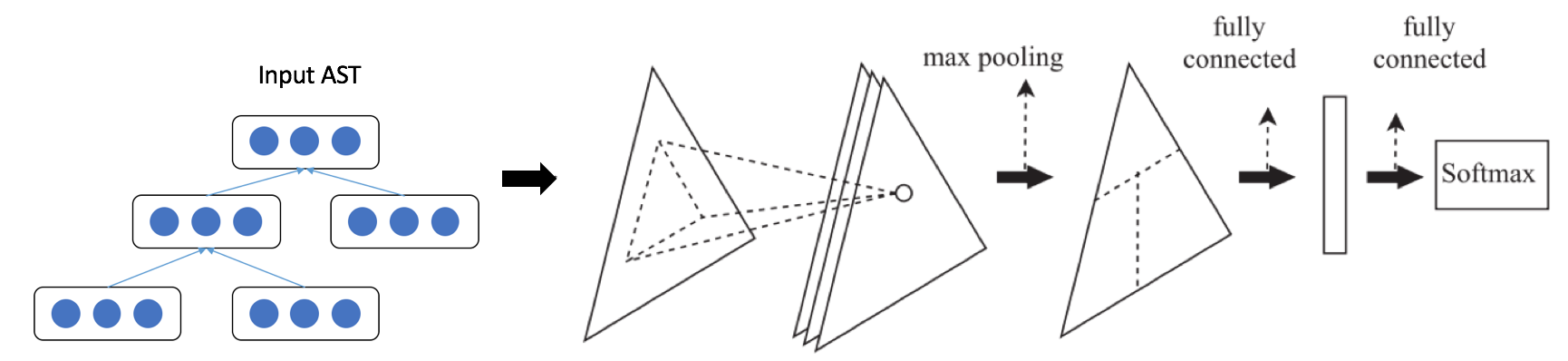}	
	\caption{TBCNNs, excerpt from~\cite{DBLP:conf/aaai/MouLZWJ16}}
	\label{figure:tbcnn}
\end{figure}

\subsection{3.1 Tree-based convolutional neural networks}
TBCNNs were proposed by \citet{DBLP:conf/aaai/MouLZWJ16}. Figure~\ref{figure:tbcnn} shows its architecture. Each AST node is represented as a vector by using an encoding layer, whose task is to embed AST node types in a continuous vector space where semantically similar types are mapped to nearby points. For examples, the types `while' and `for' are similar because they are both loop statements. We use a different strategy to embed AST node types, as described in Section 4.2.

\citet{DBLP:conf/aaai/MouLZWJ16} designed a set of fixed-depth subtree filters sliding over an entire AST to extract structural information of the tree. The pooling layer was added to gather the extracted information over various parts of the tree. They also proposed ``continuous binary trees'' and applied dynamic pooling \cite{DBLP:conf/nips/SocherHPNM11} to deal with varying numbers of children of AST nodes. Finally, they added a hidden layer and an output layer to classify programs.

\subsection{3.2 The pre-trained vector}
To train the TBCNNs, one needs an initial vector representation for each tree node. \citet{DBLP:conf/aaai/MouLZWJ16} use the ``coding criterion'' from \citet{PengMLLZJ15} to learn the vector representation for each AST node. We use a different strategy, similar to the skip-gram model of {\em word2vec} \cite{DBLP:journals/corr/abs-1301-3781}, but applied to the context of ASTs. The skip-gram model, given an input word in a sentence, looks at the words nearby and picks one at random. The model predicts the probability for each word in the whole vocabulary to be a ``nearby word'' of the input word. So the task here is to ``predict the contextual words given an input word''.

With this idea in mind, we apply it for the so-called \textit{AST2vec} task. That is, we pick random children of a given AST node. The networks, in this case, will tell us the probability for each node in the whole AST vocabulary to be the ``chosen children''. The vocabulary words, in this case, are the AST node types, which are of rather small sizes (around 450 combining C/C++, C\#, Objective C, and Java).

\subsection{3.3 The neural network classification model}
\begin{figure}[t!]\centering
	\includegraphics[width=0.8\columnwidth]{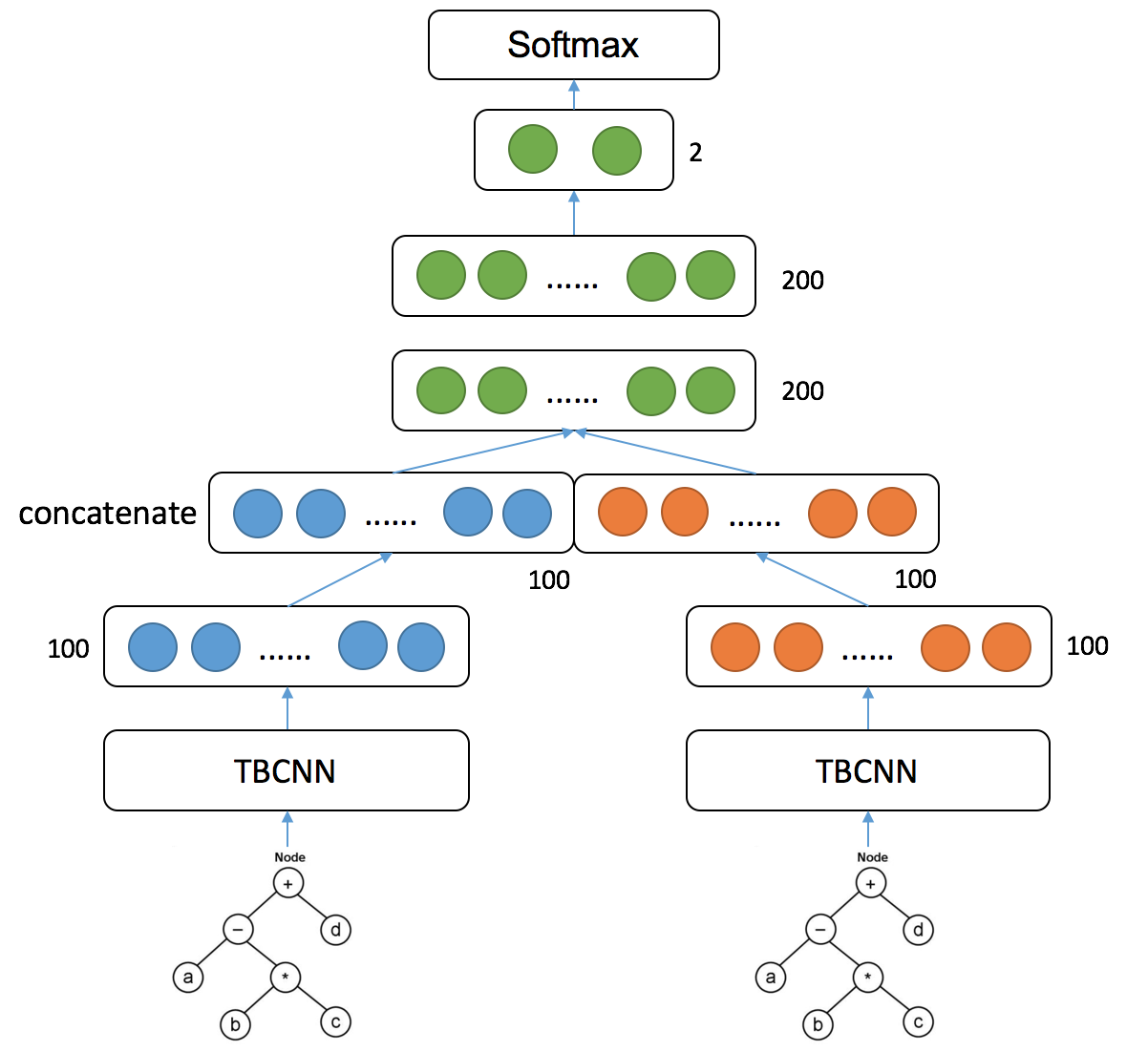}	
	\caption{BiTBCNN architecture for program classification}
	\label{figure:model}
\end{figure}

Our model in Figure~\ref{figure:model} has (1) a bilateral structure with two subnetworks, each of which processes a tree representation in parallel, and (2) the classification networks, which are simply fully connected layers connecting the two trees to the final Softmax layer, classifying if the two code snippets represented by the trees implement a same algorithm.

The subnetworks are adapted from TBCNNs~\cite{DBLP:conf/aaai/MouLZWJ16}. Each subnetwork receives the AST representation of a program as the input. The TBCNNs will perform a convolutional step to extract features from the trees. In our case, after the pooling layer of each TBCNN, we get the feature representation vector of each program, and concatenate the two vectors to a merged vector, so called the ``joint feature representation layers''. Then, two more fully connected hidden layers above the joint feature representation layers are added and connected to a Softmax layer to classify if the two input programs implement a same algorithm or not.

\section{4. Evaluation}\label{sec:evaluation}
\subsection{4.1 Datasets}
We have collected data from GitHub for six algorithms: mergesort (ms), bubblesort (bs), quicksort (qs), linkedlist (ll), breadth first search (bfs) and knapsack (kns), both in C++ and Java. For each language, we get approximately 3500 programs. The details of our dataset are depicted in Table~\ref{tab:dataset}, where the number of instances of programs crawled from the GitHub for specific algorithms are shown.

\begin{table}[t]
\centering
{\small
\caption{C++ and Java code sampled from GitHub}\label{tab:dataset}
	\begin{tabular}{c||c|c|c|c|c|c|c}
        Lang./Algo. & ms & bs & qs & ll & bfs & kns & total\\
    	\hline
        \hline
        C++ & 588 & 531 & 567 & 609 & 609 & 630 & 3534 \\
        \hline
        Java & 588 & 609 & 567 & 588 & 609 & 630 & 3591\\
    \end{tabular}
}
\end{table}


\subsection{4.2 Experiments}

Our experiments include two settings. The first checks whether BiTBCNNs perform well in classifying whether a random pair of cross-language programs implements a same algorithm. The second is to check whether the classifier still works well when it is applied to classify which algorithm an unknown program implements, based on a set of known programs in another language, simulating a cross-language learning situation that motivates this work in Section 1.

\subsubsection{4.2.1 Binary classification.}
This task is designed to verify if our model can successfully detect whether two programs from two different languages are the same or not. For programs in the original dataset of each language and each algorithm, we split 70 percent for training and 30 percent for testing. Thus we have approximately 2,500 programs on each language for training and 1,000 programs for testing. With 2,500 programs on each side, we get 6,250,000 pairs of programs (about 1,100,000 similar pairs and 5,100,000 dissimilar pairs). At this moment, we feed into the left subnetwork C++ programs and the right subnetwork Java programs. For each training epoch, we randomly select 1,000 similar pair and 1,000 dissimilar pairs to get balanced inputs for the epoch.
We train the model for 100 epochs.

For the testing, as we have approximately 1,000 C++ programs and 1,000 Java programs, we could have approximately 1,000,000 pairs in total. 
To save time, we randomly select 2,000 similar pairs and 2,000 dissimilar pairs, which amount to around 0.4\% of all the testing pairs. We use precision, recall and f1 score as the metrics to evaluate this task. The result is shown in Table \ref{tab:binary}.

\begin{table}
\centering
{\small
\caption{Results of cross-language program classifications. The first column shows the labels of pairs, 1 means similar, 0 means dissimilar. The rest are  metrics to evaluate this task.}
    \label{tab:binary}
	\begin{tabular}{c||c|c|c}
        label & precision & recall & f1 \\
        \hline\hline
        1 & 0.98 & 0.91 & 0.95 \\
        \hline
        0 & 0.92 & 0.94 & 0.93 \\
    \end{tabular}
}
\end{table}

\subsubsection{4.2.2 Algorithm Detection.}
This task evaluates how well our model performs in classifying the actual algorithm implemented by a given input program. Taking a random program A for testing, we use it as the input for the left subnetwork, and pick a known program B implementing a known algorithm and use it for the right. In this way, one can infer the algorithm label of the program A based on outputs from the above binary classifier. Note that in our experiment, we always assume that the left input is a C++ program and the right input is a Java program.

We thus take 1,000 random C++ programs from the testing data. Then for each of the C++ programs, we randomly pick one known/training Java programs from each of the six algorithm labels. We compare each C++ program with each of six Java programs using BiTBCNNs, in order to tell which one yields the highest probability in the softmax layer, and we use the algorithm label of the Java program that yields the highest probability as the predicted label of the C++ program. Finally, we compare the true label of the C++ program with the predicted one, and get a precision of 80.5\%.

\subsubsection{4.2.3 Threats to Validity.}
We have not looked at all available programming languages or algorithms.
We will need to verify whether the current code collected via GitHub search APIs may have biases,
and evaluate our approach with more languages and more algorithms, e.g., using Rosetta Code ({\scriptsize \url{http://rosettacode.org/wiki/Rosetta_Code}}).

The programs we used are relatively small with relatively clearly defined algorithms. If a program becomes larger or contains mixed set of algorithms, our approach may not be applicable directly.
Training speed may become a concern too when more data is used, although each round of training for our limited dataset only took tens of minutes on a commodity desktop machine. We think traditional program analysis (e.g., dependence-based slicing ) may be useful for alleviating such problems by partitioning a large program into smaller ones first before applying our approach.

The architecture of our BiTBCNNs may be varied in many ways as studies in the area of natural language processing have shown. And we have only used simple data dropout rate of 0.7 to reduce over-fitting. There is still much work to explore various neural network structures.
Also, our encoding of the trees removed identifier names. In future we will consider leveraging on the similarity in names and more code semantics (e.g., dependencies among code elements) for more accurate code encoding.


\section{5. Related Work}\label{sec:related}
For the problem of cross-language program translation, much work has utilized various statistical language models for tokens~\cite{Nguyen2013}, phrases~\cite{Nguyen2015,Nguyen2016}, or APIs~\cite{Zhong2013,Nguyen2014,Nguyen2014a,Phan2017,Zhong2010}. Only a few studies have used deep learning for language recognition and translation, at the API level~\cite{Gu2017,Gu2016}, which is still far from classifying functionally similar code fragment or performing translation for any code fragment.
Although some practical tools exist for translating code among specific languages (e.g., Java2CSharp: {\scriptsize \url{https://github.com/codejuicer/java2csharp}}), they are mostly rule-based, rather than statistics-based~\cite{Karaivanov:2014:PST:2661136.2661148} depending on clearly defined grammars of individual languages, and not easily extensible for different languages.

For natural languages, many studies on sentence comparisons and translations involve variants of bilateral structures as shown by \citet{wang2017bilateral}.
Among them, \citet{Bromley1993} pioneered ``Siamese'' structures to join two subnetworks for written signature comparison. \citet{He2015MultiPerspectiveSS} also use such structures to compute sentence features at multiple levels of granularity. However, these studies have not considered tree-based structures that are more accurate representations of code.

In code learning, \citet{hellendoorn2017deep} point out that simpler models (e.g., n-gram) improved with cached information about code locality and hierarchy may even outperform complex models (e.g., deep neural networks). But this also indicates to us that incorporating code locality and structural information with deep learning by using tree-based convolutional neural networks (TBCNNs) may improve code learning accuracy. Using TreeNNs, \citet{DBLP:conf/icml/AllamanisCKS17} propose to represent symbolic expressions; however, it has not been applied to other type of code structures. Although \citet{DBLP:conf/aaai/MouLZWJ16} introduce TBCNNs to classify C++ programs based on functionality and to detect code of certain patterns and others use tree-based encodings too (e.g., \citet{white2016deep} for code clone detection), it has not been applied to cross-language program classification.

\section{6. Conclusions \& Future Work}\label{sec:conclusion}
In this paper, we have presented the BiTBCNNs approach to the cross-language program classification problem, where algorithms are identified from source AST structures automatically. Using benchmarks of algorithms crawled from GitHub, we have shown that it is possible to train a model on multiple languages, with an accuracy of above 80\%. The number and representativeness of training datasets may affect the ultimate performance, while cross-language deep learning makes it likely possible to reuse the implementation of algorithms from different languages.

Our future work include tuning BiTBCNNs structures and parameters, supporting more programming languages and more algorithms with more training data, learning from more code semantic information such as dependence data, and applying to more tasks such as cross-language code clone detection, algorithm patent protection, and bug fixing.

\section*{Acknowledgements}
This work is supported by eSTEeM project on Ask Programming Aloud and EPSRC Platform Grant on SAUSE.


\let\oldthebibliography=\thebibliography
\let\endoldthebibliography=\endthebibliography
\renewenvironment{thebibliography}[1]{%
  \begin{oldthebibliography}{#1}%
    \setlength{\parskip}{0ex}%
    \setlength{\itemsep}{1ex}%
}%
{%
  \end{oldthebibliography}%
}
\tiny
\bibliography{main}

\begin{thebibliography}{}

\bibitem[\protect\citeauthoryear{Bowman \bgroup et al\mbox.\egroup
  }{2015}]{DBLP:journals/corr/BowmanAPM15}
Bowman, S.~R.; Angeli, G.; Potts, C.; and Manning, C.~D.
\newblock 2015.
\newblock A large annotated corpus for learning natural language inference.
\newblock {\em CoRR} abs/1508.05326.

\bibitem[\protect\citeauthoryear{Gu \bgroup et al\mbox.\egroup }{2016}]{Gu2016}
Gu, X.; Zhang, H.; Zhang, D.; and Kim, S.
\newblock 2016.
\newblock Deep {API} learning.
\newblock In {\em Proceedings of the 24th {ACM} {SIGSOFT} International
  Symposium on Foundations of Software Engineering (FSE)},  631--642.

\bibitem[\protect\citeauthoryear{Gu \bgroup et al\mbox.\egroup }{2017}]{Gu2017}
Gu, X.; Zhang, H.; Zhang, D.; and Kim, S.
\newblock 2017.
\newblock {DeepAM}: Migrate {APIs} with multi-modal sequence to sequence
  learning.
\newblock In {\em Proceedings of the Twenty-Sixth International Joint
  Conference on Artificial Intelligence (IJCAI)},  3675--3681.

\bibitem[\protect\citeauthoryear{He, Gimpel, and
  Lin}{2015}]{He2015MultiPerspectiveSS}
He, H.; Gimpel, K.; and Lin, J.~J.
\newblock 2015.
\newblock Multi-perspective sentence similarity modeling with convolutional
  neural networks.
\newblock In {\em EMNLP}.

\bibitem[\protect\citeauthoryear{Hellendoorn and
  Devanbu}{2017}]{hellendoorn2017deep}
Hellendoorn, V.~J., and Devanbu, P.
\newblock 2017.
\newblock Are deep neural networks the best choice for modeling source code?
\newblock In {\em Proceedings of the 2017 11th Joint Meeting on Foundations of
  Software Engineering, ser. ESEC/FSE},  763--773.

\bibitem[\protect\citeauthoryear{Mikolov \bgroup et al\mbox.\egroup
  }{2013}]{DBLP:journals/corr/abs-1301-3781}
Mikolov, T.; Chen, K.; Corrado, G.; and Dean, J.
\newblock 2013.
\newblock Efficient estimation of word representations in vector space.
\newblock {\em CoRR} abs/1301.3781.

\bibitem[\protect\citeauthoryear{Mou \bgroup et al\mbox.\egroup
  }{2014}]{DBLP:journals/corr/MouLLPJX014}
Mou, L.; Li, G.; Liu, Y.; Peng, H.; Jin, Z.; Xu, Y.; and Zhang, L.
\newblock 2014.
\newblock Building program vector representations for deep learning.
\newblock {\em CoRR} abs/1409.3358.

\bibitem[\protect\citeauthoryear{Mou \bgroup et al\mbox.\egroup
  }{2016}]{DBLP:conf/aaai/MouLZWJ16}
Mou, L.; Li, G.; Zhang, L.; Wang, T.; and Jin, Z.
\newblock 2016.
\newblock Convolutional neural networks over tree structures for programming
  language processing.
\newblock In {\em Proceedings of the Thirtieth {AAAI} Conference on Artificial
  Intelligence},  1287--1293.

\bibitem[\protect\citeauthoryear{Nguyen \bgroup et al\mbox.\egroup
  }{2014a}]{Nguyen2014a}
Nguyen, A.~T.; Nguyen, H.~A.; Nguyen, T.~T.; and Nguyen, T.~N.
\newblock 2014a.
\newblock Statistical learning approach for mining {API} usage mappings for
  code migration.
\newblock In {\em {ACM/IEEE} International Conference on Automated Software
  Engineering (ASE)},  457--468.

\bibitem[\protect\citeauthoryear{Nguyen \bgroup et al\mbox.\egroup
  }{2014b}]{Nguyen2014}
Nguyen, A.~T.; Nguyen, H.~A.; Nguyen, T.~T.; and Nguyen, T.~N.
\newblock 2014b.
\newblock Statistical learning of {API} mappings for language migration.
\newblock In {\em Proceedings of the 36th International Conference on Software
  Engineering - Companion (ICSE)},  618--619.

\bibitem[\protect\citeauthoryear{Nguyen, Nguyen, and Nguyen}{2013}]{Nguyen2013}
Nguyen, A.~T.; Nguyen, T.~T.; and Nguyen, T.~N.
\newblock 2013.
\newblock Lexical statistical machine translation for language migration.
\newblock In {\em Joint Meeting of the European Software Engineering Conference
  and the {ACM} {SIGSOFT} Symposium on the Foundations of Software Engineering
  (ESEC/FSE)},  651--654.

\bibitem[\protect\citeauthoryear{Nguyen, Nguyen, and Nguyen}{2015}]{Nguyen2015}
Nguyen, A.~T.; Nguyen, T.~T.; and Nguyen, T.~N.
\newblock 2015.
\newblock Divide-and-conquer approach for multi-phase statistical migration for
  source code {(T)}.
\newblock In {\em 30th {IEEE/ACM} International Conference on Automated
  Software Engineering (ASE)},  585--596.

\bibitem[\protect\citeauthoryear{Nguyen, Tu, and Nguyen}{2016}]{Nguyen2016}
Nguyen, A.~T.; Tu, Z.; and Nguyen, T.~N.
\newblock 2016.
\newblock Do contexts help in phrase-based, statistical source code migration?
\newblock In {\em 2016 {IEEE} International Conference on Software Maintenance
  and Evolution (ICSME)},  155--165.

\bibitem[\protect\citeauthoryear{Peng \bgroup et al\mbox.\egroup
  }{2015}]{PengMLLZJ15}
Peng, H.; Mou, L.; Li, G.; Liu, Y.; Zhang, L.; and Jin, Z.
\newblock 2015.
\newblock Building program vector representations for deep learning.
\newblock In {\em Proceedings of the 8th International Conference on Knowledge
  Science, Engineering and Management (KSEM)},  547--553.

\bibitem[\protect\citeauthoryear{Phan \bgroup et al\mbox.\egroup
  }{2017}]{Phan2017}
Phan, H.~D.; Nguyen, A.~T.; Nguyen, T.~D.; and Nguyen, T.~N.
\newblock 2017.
\newblock Statistical migration of {API} usages.
\newblock In {\em Proceedings of the 39th International Conference on Software
  Engineering - Companion Volume (ICSE)},  47--50.

\bibitem[\protect\citeauthoryear{Socher \bgroup et al\mbox.\egroup
  }{2011}]{DBLP:conf/nips/SocherHPNM11}
Socher, R.; Huang, E.~H.; Pennington, J.; Ng, A.~Y.; and Manning, C.~D.
\newblock 2011.
\newblock Dynamic pooling and unfolding recursive autoencoders for paraphrase
  detection.
\newblock In {\em Proceedings of the 24th International Conference on Neural
  Information Processing Systems (NIPS)},  801--809.

\bibitem[\protect\citeauthoryear{White \bgroup et al\mbox.\egroup
  }{2016}]{white2016deep}
White, M.; Tufano, M.; Vendome, C.; and Poshyvanyk, D.
\newblock 2016.
\newblock Deep learning code fragments for code clone detection.
\newblock In {\em Proceedings of the 31st IEEE/ACM International Conference on
  Automated Software Engineering},  87--98.
\newblock ACM.

\bibitem[\protect\citeauthoryear{Zhong \bgroup et al\mbox.\egroup
  }{2010}]{Zhong2010}
Zhong, H.; Thummalapenta, S.; Xie, T.; Zhang, L.; and Wang, Q.
\newblock 2010.
\newblock Mining {API} mapping for language migration.
\newblock In {\em Proceedings of the 32nd {ACM/IEEE} International Conference
  on Software Engineering - Volume 1 (ICSE)},  195--204.

\bibitem[\protect\citeauthoryear{Zhong, Thummalapenta, and
  Xie}{2013}]{Zhong2013}
Zhong, H.; Thummalapenta, S.; and Xie, T.
\newblock 2013.
\newblock Exposing behavioral differences in cross-language {API} mapping
  relations.
\newblock In {\em Proceedings of 16th International Conference on Fundamental
  Approaches to Software Engineering (FASE), Held as Part of the European Joint
  Conferences on Theory and Practice of Software (ETAPS)},  130--145.

\end{thebibliography}
\bibliographystyle{aaai}

\end{document}